\documentclass{article}

\usepackage{microtype}
\usepackage{graphicx}
\usepackage{subfigure}
\usepackage{booktabs} 
\usepackage{hyperref}

\usepackage[accepted]{style_class}

\usepackage{amsmath}
\usepackage{amssymb}
\usepackage{mathtools}
\usepackage{amsthm}

\usepackage[capitalize,noabbrev]{cleveref}

\theoremstyle{plain}

\theoremstyle{definition}

\theoremstyle{remark}

\DeclareMathOperator*{\argmax}{argmax} 
\DeclareMathOperator{\proj}{proj} 
\DeclareMathOperator{\ident}{id} 

\newcommand{\resMSE}{\textnormal{ResMSE}}
\newcommand{\diff}{{\,\mathrm{d}}}

\newcommand{\eps}{\varepsilon}

\newcommand{\Nc}{\mathcal{N}}
\newcommand{\Sc}{\mathcal{S}}
\newcommand{\Tc}{\mathcal{T}}
\newcommand{\Uc}{\mathcal{U}}

\newcommand{\N}{\mathbb{N}}
\newcommand{\R}{\mathbb{R}}

\usepackage[textsize=tiny]{todonotes}

\icmltitlerunning{General Transform-Invariant PCA}

\begin{document}

\twocolumn[
\icmltitle{GT-PCA: Effective and Interpretable Dimensionality Reduction with General Transform-Invariant Principal Component Analysis}



\icmlsetsymbol{equal}{*}

\begin{icmlauthorlist}
\icmlauthor{Florian Heinrichs}{yyy,zzz}
\end{icmlauthorlist}

\icmlaffiliation{yyy}{Department of Computer Science, Darmstadt University of Applied Sciences,  Germany}
\icmlaffiliation{zzz}{Hessian Center for Artificial Intelligence (hessian.AI), Darmstadt, Germany}

\icmlcorrespondingauthor{Florian Heinrichs}{florian.heinrichs@h-da.de}


\vskip 0.3in
]



\printAffiliationsAndNotice{}  

\begin{abstract}
	Data analysis often requires methods that are invariant with respect to specific transformations, such as rotations in case of images or shifts in case of images and time series. While principal component analysis (PCA) is a widely-used dimension reduction technique, it lacks robustness with respect to these transformations. Modern alternatives, such as autoencoders, can be invariant with respect to specific transformations but are generally not interpretable. We introduce General Transform-Invariant Principal Component Analysis (GT-PCA) as an effective and interpretable alternative to PCA and autoencoders. We propose a neural network that efficiently estimates the components and show that GT-PCA significantly outperforms alternative methods in experiments based on synthetic and real data.
\medskip


\end{abstract}

\section{Introduction} \label{sec:intro}

Over the last years, the volume of collected data was steadily increasing. This effect is due to the expanding use of sensors and due to the higher resolution with which data is measured. With larger amounts of data, the need for effective dimension reduction techniques is growing as well. On the one hand, data analysis becomes increasingly difficult in higher dimensions because of the \textit{curse of dimensionality}. On the other hand, effective methods for dimension reduction are relevant also outside of statistics and machine learning, for example to compress and store data efficiently.

\begin{figure}[tb]
	\begin{center}
		\includegraphics[width=0.45\columnwidth]{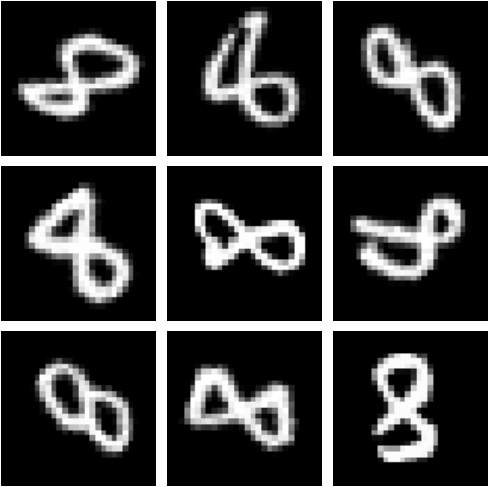}
		\hspace{0.03\columnwidth}
		\includegraphics[width=0.45\columnwidth]{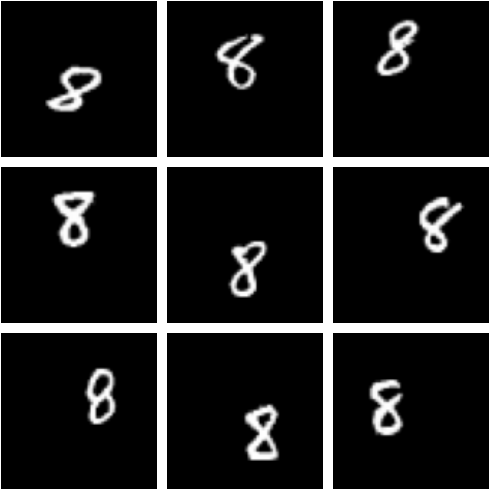}
		\caption{Exemplary transformed MNIST data. Left: rotated digits. Right: shifted digits.}
		\label{fig:mnist}
	\end{center}
	\vskip -0.2in
\end{figure}

When reducing the dimension of data, we try to keep relevant information while discarding non-relevant details. Often, only the shape of an object is of interest. For instance, when recognizing handwritten digits, as displayed in Figure \ref{fig:mnist}, we are primarily interested in the digit's value rather than its position or orientation. 

In real-time applications based on sensor data, we often work with ``sliding windows'', where a signal of interest might occur at any time in the window. In this case, we need methodology that is shift-invariant in the sense of detecting a signal independently of its position in a window. 

More generally, we are often interested only in specific features of an object that are stable with respect to certain transformations, such as rotations and reflections of images, or shifts of time series. 

A traditional approach to dimension reduction is principal component analysis (PCA). PCA and its extensions, like functional PCA for functional data, require the data to be aligned, i.\,e., the signal of interest must have the same position in each observation. This assumption is often violated for data such as images or windows of time series. Aligning all observations is often infeasible, especially if the signal at the border of an observation is only partially contained or if multiple signals are present at different time points.

There are various extensions of PCA that are, to some extent, invariant with respect to certain transformations. Recently, autoencoders have become particularly popular. They were proposed in 1991 as nonlinear PCA and consist of a symmetric neural network with a "bottleneck" layer in the middle \cite{kramer1991}. However, as most neural networks, autoencoders are generally not interpretable.

In this work, we propose general transform-invariant principal components analysis (GT-PCA). The proposed components are invariant w.r.t. problem-dependent transformations and can be easily interpreted. Thus, GT-PCA combines the advantages of PCA with those of modern autoencoders. 

We summarize our contribution as follows:\vspace{-0.1in}
\begin{itemize}
	\item We propose GT-PCA, an extension of PCA, that is invariant w.r.t problem-dependent transformations.
	\item We present a neural network to approximate the proposed general transform-invariant principal components (GT-PCs).\footnote{Python code available at \url{https://github.com/FlorianHeinrichs/gt_pca}.}
	\item We show that GT-PCA effectively reduces the dimension of synthetic and real data and successfully use the lower dimensional projections for subsequent classification and outlier detection.
\end{itemize}
\vspace{-0.1in}

Due to the general formulation of GT-PCs, the proposed methodology can be used in a variety of domains, from the analysis of images and videos to brain activity, measured via electroencephalography, and essentially any type of data that can be modeled as function of one or more variables.


\section{Related Work} \label{sec:rel_work}

To the best of our knowledge, transform-invariant PCA is explicitly discussed only for the special case of face alignment \cite{deng2013}. However, a large variety of nonlinear PCA extensions exist in the literature, some of which are invariant w.r.t. certain transformations. 

Schölkopf et al. proposed kernel PCA (KPCA), a variant of PCA that uses a nonlinear function to map data into some high-dimensional space \yrcite{scholkopf1997}. Hastie \& Stuetzle proposed to project high dimensional points onto a smooth one-dimensional curve, referred to as principal curve \yrcite{hastie1989}. Unlike principal components, principle curves do not yield ``loadings'' of projected data points, which makes dimension reduction difficult. Hence, Dong \& McAvoy proposed to combine principal curves with neural networks that yield loadings \yrcite{dong1996}.
 
Further, PCA has been extended to functional data \citep{ramsay2005}. Song \& Li proposed nonlinear and additive PCA, which is essentially an extension of KPCA to functional data \yrcite{song2021}. 

More recently, significant advances in dimension reduction have been made through the use of deep learning. Kramer introduced nonlinear PCA based on neural networks, which is nowadays referred to as autoencoders \yrcite{kramer1991}. These autoencoders (AEs) exist in a variety of flavors, such as convolutional AEs \cite{makhzani2015, holden2015}, variational AEs (VAEs) \citep{Kingma2014} and convolutional VAEs \citep{kulkarni2015}. Kanavati et al. proposed $\sigma$-PCA, a neural network that aims to combine the benefits of linear and nonlinear PCA \yrcite{kanavati2023}. As with all neural networks, the interpretation of trained models remains an open problem. 

A possible remedy is the sequential training of components, as proposed in the initial work by Kramer \yrcite{kramer1991}. However, in this work the use of fully connected layers was proposed, which does not yield the desired transform-invariance. We are taking up this idea and sequentially estimate the proposed GT-PCs through adequate layers, depending on the problem-dependent transformations.

\section{Methodology} 
\label{sec:main}

In this section, we introduce \textit{general transform-invariant principal component analysis} (GT-PCA) and discuss special cases of transformations and their corresponding data types. In contrast to PCA, it is infeasible to calculate the proposed components analytically. Therefore, we present a neural network architecture that approximates these components.

\subsection{General Transform-Invariant Principal Components Analysis}\label{sec:gt_pc}

Assume we have observations $X_1, \dots, X_n$ in a Hilbert space $H$, with inner product $\langle \cdot,\cdot \rangle$ and norm $\|\cdot\|$. Important examples are $H=\R^d$ and $H=L^2([0,1])$. Further, let $\Tc$ be a family of invertible linear operators. We are interested in principal components (PCs) that are invariant w.r.t. transformations $T\in\Tc$. 

Recall the definition of the conventional first PC 
$$ w_1 = \argmax_{w \in H} \sum_{i=1}^{N} \frac{\langle X_i, w \rangle^2}{\|w\|^2}. $$%

Similarly, we define the first $\Tc$-invariant GT-PC as 
\begin{equation*}
	v_1 = \argmax_{w \in H} \sum_{i=1}^{N} \max_{T\in\Tc} \frac{\langle X_i, Tw \rangle^2}{\|Tw\|^2}. 
\end{equation*} 

Intuitively, the inner maximum finds the transform of $w$ that best aligns with each sample $X_i$. Analogously to conventional PCs, we define the succeeding components as%
\begin{equation}\label{eq:def_components}
	v_k = \argmax_{w \in H} \sum_{i=1}^{N} \max_{T\in\Tc} \frac{\langle \hat{X}_i^{(k)}, Tw \rangle^2}{\|Tw\|^2} 
\end{equation} 
based on the residuals $\hat{X}_i^{(k)} = X_i - \proj_{k-1}(X_i)$ and the projections
\begin{equation}\label{eq:def_projections}
	\proj_{k-1}(X_i) = \sum_{j=1}^{k-1} \max_{T\in\Tc} \frac{\langle \hat{X}_i^{(j)}, Tv_j \rangle}{\|Tv_j\|} Tv_j.
\end{equation}

If $\Tc = \{\ident\}$, we obtain conventional PCA for $H=\R^d$ and functional PCA for $H=L^2([0,1])$. Contrarily, if $\Tc = \{T:H\to H| T~ \textnormal{linear and invertible}\}$, the components will be invariant w.r.t. any invertible linear transformation, which renders them useless. Thus, the choice of the operator family $\Tc$ is crucial to obtain a meaningful generalization of PCA. Depending on the type of data, different transformations are commonly used, as explained below.

\subsubsection{Time-dependent Data} \label{sec:time_dep_data}

Assume that we have time-dependent data $X_1(t), \dots, X_n(t)$, for time $t \in [0, 1]$, where some signal of interest is located at any position in the interval $[0, 1]$. In this case, we want the PCA to be invariant w.r.t. a time shift. Hence, we can define $\Tc = \{ T_s\}_{s\in \Sc}$ as the set of all shift operators $T_s$ mapping a function $w$ to $w(\cdot-s)$, where $\Sc\subset\R$ denotes an index set.

This definition also allows for signals $w:[0, h]\to\R$ with a different length compared to the data. For smaller signals with $h\in(0,1)$, we define $\Sc=[0, 1-h]$ and set $w$ to zero outside of its domain $[0,h]$. For larger signals with $h>1$, we define $\Sc=[-h+1, 0]$. In both cases, the shifted functions $T_sw$ are properly defined on $[0, 1]$. If the data and weights are in $L^2([0, 1])$, we can use the inner product $\langle f, g\rangle = \int_0^1 f(x)g(x)\diff x$ and its induced norm $\|f\|^2 = \sqrt{\langle f, f \rangle}$ to define shift-invariant PCA.

\begin{figure}[t]
	\begin{center}
		\includegraphics[width=\columnwidth]{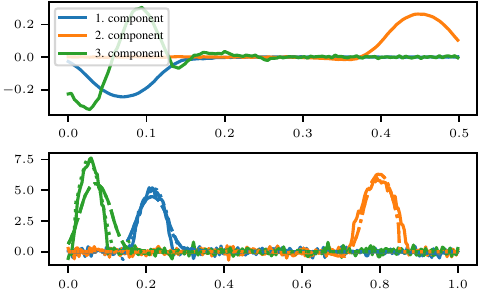}%
		\caption{Top: First 3 GT-PCs of generated spikes with window length $\tfrac{1}{2}$. Bottom: Generated signals of length $1$ with spikes at different times, different widths and amplitudes and their reconstructions based on 1 (dashed), 2 (dashdot) and 3 (dotted) components.}
		\label{fig:spikes_greater}
	\end{center}
	\vskip -0.2in
\end{figure}

\begin{figure}[!t]
	\vskip 0.1in
	\begin{center}
		\includegraphics[width=\columnwidth]{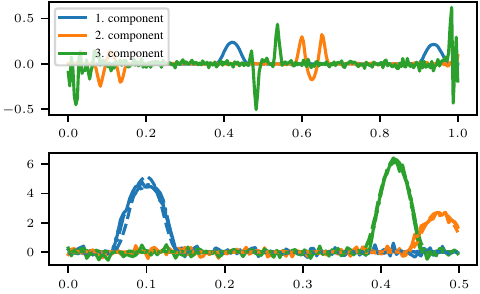}%
		\caption{Top: First 3 GT-PCs of generated spikes with window length $1$. Bottom: Generated signals of length $\tfrac{1}{2}$ with spikes at different times, different widths and amplitudes and their reconstructions based on 1 (dashed), 2 (dashdot) and 3 (dotted) components.}
		\label{fig:spikes_less}
	\end{center}
	\vskip -0.2in
\end{figure}

Exemplary data of this type on the intervals $[0, 1]$ and $[0, 0.5]$ is displayed on the bottom of Figures \ref{fig:spikes_greater} and \ref{fig:spikes_less}. The data generating process is described in Section \ref{sec:sim_study}. The corresponding first three components on the intervals $[0, 0.5]$ and $[0, 1]$ are displayed on the bottom of the respective figures. The three components have characteristic peaks of different widths, which allows the modeling of differently shaped signals. The signals can be effectively projected onto a single component without losing much information. The resulting GT-PCA for functional data can be interpreted similar to functional PCA, as described by Ramsay \& Silverman \yrcite{ramsay2005}.

If the data is not given as functions of continuous time, but sampled discretely, thus yielding time series $\{X_1^{(i)}\}_{i\in\N}, \dots,  \{X_n^{(i)}\}_{i\in\N}$, we can analogously define shift-invariant PCA based on discretized weights. 

Clearly, these shift-invariant principal components are not only useful for time-dependent data, but more generally for data depending on a single variable.


\subsubsection{Images}\label{sec:img_data}

Another important class of data are images, which can be considered as functions $f$ of two coordinates $x$ and $y$, where the function value is either in $\R$, for grayscale images, or in $\R^3$ for RGB images. 

In this context, different transformations are of interest, e.\,g., reflection operators $T_\sigma(f)(x, y)=f(\sigma_1 x, \sigma_2 y)$, for $\sigma \in \{-1, 1\}^2$, or scaling operators $T_s(f)(x, y) = f(sx, sy)$, for $s>0$. Another important class of transformations for images are rotations of the form %
$$T_\alpha(f)(x, y) = f\big(\cos(\alpha) x - \sin(\alpha) y, \sin(\alpha) x + \cos(\alpha) y\big).$$%

\begin{figure}[tb]
	\begin{center}
		\includegraphics[width=0.45\columnwidth]{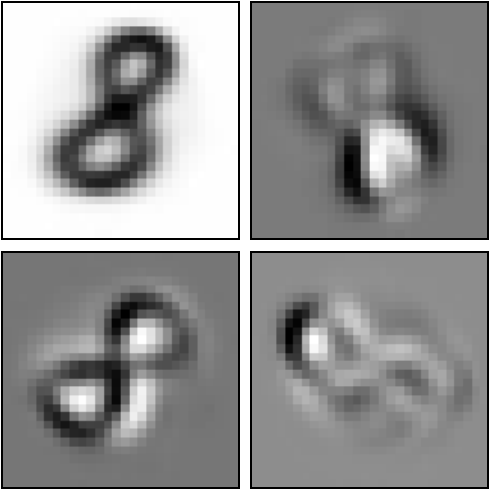}
		\hspace{0.03\columnwidth}
		\includegraphics[width=0.45\columnwidth]{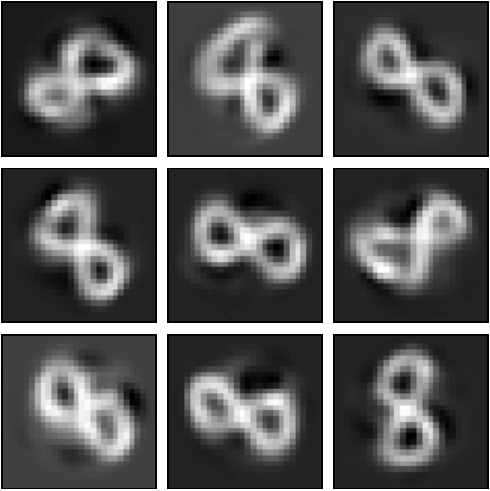}\\
		\vspace{0.1in}
		\hspace{0.001\columnwidth}
		\includegraphics[width=0.45\columnwidth]{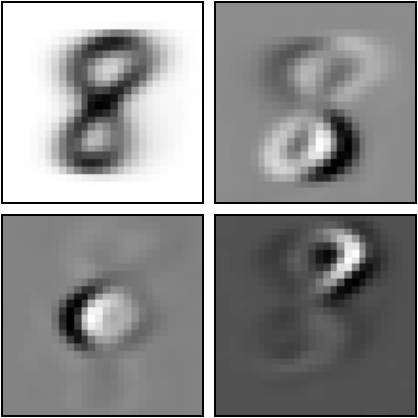}
		\hspace{0.03\columnwidth}
		\includegraphics[width=0.45\columnwidth]{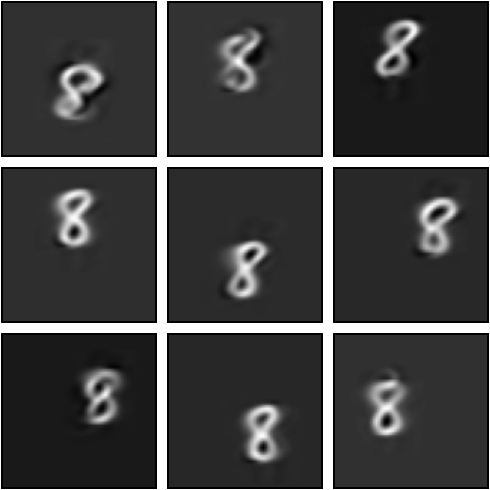}
		\caption{Top left: First four GT-PCs of rotated digits. Top right: Reconstruction of rotated digits based on first four GT-PCs. Bottom left: First four GT-PCs of shifted digits. Bottom right: Reconstruction of shifted digits based on first four GT-PCs.}
		\label{fig:mnist_shift}
	\end{center}
	\vskip -0.2in
\end{figure}

On the left of Figure \ref{fig:mnist}, nine randomly rotated handwritten digits from the MNIST data set are displayed \citep{lecun1998}. On the top of Figure \ref{fig:mnist_shift}, the first four components and reconstructions of the rotated digits based on these components are visualized. It can be seen that with as little as four GT-PCs, the characteristic shape and orientation of different digits can be restored effectively. 

Finally, we can define shift-invariant principle components based on the operators $T_{u, v}(w)(x, y) = w(x-u, y-v)$ for appropriate index sets $U$ and $V$ in $\R$. The same remarks about different sizes of the weights as in the one-dimensional case apply, so that GT-PCA for data in the space $L^2([0, 1]^2)$ can be properly defined.

Examples of handwritten digits at different locations in space are displayed on the right of Figure \ref{fig:mnist}, and their first four components and reconstructions based on these components are visualized on the bottom of Figure \ref{fig:mnist_shift}. As with the rotated digits, the characteristic shape and position in space can be restored effectively. 

Again, the components can be interpreted similarly to functional PCA. Whereas the first component corresponds to the rescaled mean of the data, the other  components are responsible for changing the digits' shapes. The second and fourth components primarily impact the bottom and top of the digits, respectively.

As before, when data is not sampled continuously, we simply replace the continuous transformations by their discrete counterparts.

\subsection{Neural Network-based Approximation of GT-PCs}

Solving the optimization problem, that defines the GT-PCs, is infeasible for many interesting families of transformations. Hence, we propose to approximate GT-PCs in terms of a neural network with an adequate structure. 

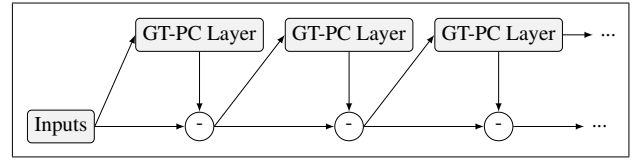
\begin{figure}[tb]
	\begin{center}
		\resizebox{\columnwidth}{!}{\frame{
\begin{tikzpicture}
		
	\node[draw, fill=black!5!white, rounded corners=2] (input) {Inputs};
	\node[circle, draw] (diff1) [right=1.5cm of input] {-};
	\node[circle, draw] (diff2) [right=2cm of diff1] {-};
	\node[circle, draw] (diff3) [right=2cm of diff2] {-};
	\node[draw, fill=black!5!white, rounded corners=2] (pc1) [above=of diff1] {GT-PC Layer};
	\node[draw, fill=black!5!white, rounded corners=2] (pc2) [above=of diff2] {GT-PC Layer};
	\node[draw, fill=black!5!white, rounded corners=2] (pc3) [above=of diff3] {GT-PC Layer};

	\draw[-latex] (input.east) -- (diff1.west);
	\draw[-latex] (diff1.east) -- (diff2.west);
	\draw[-latex] (diff2.east) -- (diff3.west);
	
	\draw[-latex] (input.east) -- (pc1.west);
	\draw[-latex] (diff1.east) -- (pc2.west);
	\draw[-latex] (diff2.east) -- (pc3.west);
	
	\draw[-latex] (pc1.south) -- (diff1.north);
	\draw[-latex] (pc2.south) -- (diff2.north);
	\draw[-latex] (pc3.south) -- (diff3.north);
		
	\draw[-latex] (diff3.east) -- +(1.18,0)
	node[right]{...};
	\draw[-latex] (pc3.east) -- +(0.5,0)
	node[right]{...};
	
	\path (-0.8,-0.5) -- (8,2.05);
\end{tikzpicture}
}}%
		\caption{Neural network architecture for the approximation of general-transform invariant principal components.}
		\label{fig:neural_network}
	\end{center}
	\vskip -0.2in
\end{figure}

The central building block of this approximation is the \texttt{GT-PC}-layer. In this layer, for some input $X\in H$, the maximum 
\begin{equation}\label{eq:layer_max}
	\max_{T\in\Tc}\frac{\langle X, Tw \rangle^2}{\|Tw\|^2}
\end{equation} is calculated and then maximized w.r.t. $w$ using some form of (stochastic) gradient descent. The argmax of this maximization is the GT-PC related to the layer. The output of a \texttt{GT-PC}-layer is the projection of its input onto the associated component. According to the definition of the projections in \eqref{eq:def_projections} and the components in \eqref{eq:def_components}, we define a neural network based on the \texttt{GT-PC}-layer as in Figure \ref{fig:neural_network}, that is sequentially trained to obtain approximations of the GT-PCs. 

\section{Experimental Results}  \label{sec:emp_res}

PCA is the gold standard for dimension reduction. We compare it to the proposed GT-PCA and additionally use KPCA, autoencoders (AEs) and variational AEs (VAEs) as baseline models. We test the proposed methodology based on synthetic and real data sets. We evaluate the models' based on the reconstruction errors of their respective projections and based on the accuracies on a consecutive classification task.

For PCA and KPCA, we used the scikit-learn implementations with their default parameters \citep{scikit-learn}. 
For the encoder parts of the (V)AEs, we used 2 blocks of convolutional and max pooling layers and a final dense layer. For the decoders, we used a dense layer, 2 blocks of convolutional and upsampling layers and a final convolutional layer. The convolutions in the encoder had 16 and 8 filters and those in the decoders 8, 16 and 1 filter. The filter size was $3$ for convolutional and $2$ for pooling and upsampling layers, respectively $3\times 3$ and $2\times 2$ for image data. The bottleneck dimension of the (V)AEs, was equal to the number of components considered for the PCA-based methods. We used the Adam optimizer with its default parameters in TensorFlow to minimize the reconstruction MSE \citep{tensorflow}. 
Note that a pair of mean and variance was fitted for each component of the VAEs, thus the number of parameters is twice as large. For GT-PCA, we used the Adam optimizer to maximize the term in \eqref{eq:layer_max} w.r.t. the weight $w$.

We split the data sets into training and test data and repeated each experiment 10 times. The number of considered components depended on the complexity of the respective data set. We evaluated the quality of the projections in terms of the \textit{residual MSE} (ResMSE), which we define as 
\begin{equation*}
	\resMSE_k = \frac{\|\proj_k(X) - X \|^2}{\|X\|^2},
\end{equation*}
where $\|\cdot\|$ denotes the norm in the corresponding space. A trivial projection $\proj \equiv 0$ yields a ResMSE of 1, thus, helpful projections should yield values between 0  and 1, where smaller values indicate better projections. 

For subsequent classification and outlier detection based on the projections, we use a simple multi-layer perceptron (MLP) with a single hidden layer of 10 neurons, that is trained with the default Adam optimizer minimizing the categorical crossentropy. We evaluate the resulting predictors based on their accuracies. 

\subsection{Synthetic Data Experiments} \label{sec:sim_study}

The synthetic data sets are inspired by brain activity recorded via electroencephalography (EEG) and are adapted from Heinrichs et al. \yrcite{heinrichs2023}.

\subsubsection{Projection} \label{sec:sim_study_proj}
The first data set models the interference of different frequencies. More specifically, for each sample, we randomly chose frequencies $\alpha\sim\Uc_{[8, 12]}, \beta\sim\Uc_{[13,30]}$ and a time shift $t_0\sim \Uc_{[0, 1]}$, and defined the continuous signal
$$ f(x) = 0.6 \sin(2\pi \alpha (x + t_0)) + 0.4 \cos(2\pi \beta (x + t_0)). $$
We generated a noisy time series $X^{(t)} = f(t/T) + 0.2 \cdot \eps_t$, for $t=1, \dots, T$ and the sample frequency $T=256$, where $\{\eps_t\}_{t=1}^T$ denotes a family of independent standard normally distributed errors. 

We generated 1000 such time series each for the training and test set. For this data set, we evaluated shift-invariant PCA and used weights of lengths (i) $256$ and (ii) $512$. As explained in Section \ref{sec:gt_pc}, GT-PCs are expected to be equal to conventional PCs in case (i).

We trained the (V)AEs for 5 epochs of 500 batches with size 32. The GT-PCs were trained sequentially with 5 epochs per component. The ResMSEs are displayed in Figure \ref{fig:oscillations_results} and Table \ref{tab:oscillations_results} in the appendix.

As expected, for case (i), GT-PCs coincide with conventional PCs. The ResMSE in this case is lower for the (V)AEs compared to PCA-based methods for few components, and approximately equal for 8 and more components. The ResMSE of the KPCA is larger for all components and remains high even for 10 components.

In case (ii), the relation between (V)AEs and PCA is generally the same, as the ResMSE is lower for few components and equal for 8 and more components. The difference compared to (i) lies in the GT-PCA, that has a significantly lower ResMSE over all components and is superior even for 10 components with $\resMSE_{10}\approx 20\%$ compared to  $\resMSE_{10}\ge  40\%$ for the other models.

\begin{figure}[tb]
	\begin{center}
		\centerline{\includegraphics[width=\columnwidth]{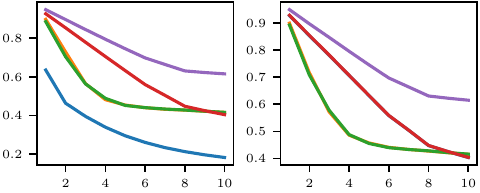}}
		\caption{Residual MSE (y-axis) for oscillating data and varying number of components (x-axis) of various models (blue: GT-PCA, red: PCA, purple: KPCA, orange: AE, green: VAE) Left: Weight length $512$. Right: Weight length $256$.}
		\label{fig:oscillations_results}
	\end{center}
	\vskip -0.2in
\end{figure}%

\subsubsection{Classification \& Outlier Detection} \label{sec:sim_study_class}

The second data set models ``spike waves'', which are important in the diagnostic analysis of EEG data. For each of the data points, we randomly chose a time $t_0\sim \Uc_{[0, 1]}$, an amplitude $\alpha\sim\Uc_{[0.5, 2.5]}$ and a spike width $w\sim\Uc_{[0.05, 0.1]}$, and defined the continuous signal
$$ f(x) = \alpha \cdot \max\big\{ -\tfrac{4}{w^2}(x-t)^2 + 3, 0 \big\}. $$
We generated a noisy time series $X^{(t)} = f(t/T) + 0.2 \cdot \eps_t$, for $t=1, \dots, T$ and some sample frequency $T$, where $\{\eps_t\}_{t=1}^T$ again denotes a family of independent standard normally distributed errors. 

Again, we generated 1000 such time series each for the training and test set and evaluated shift-invariant PCA. For $w_\ell$ denoting the weights' length, we had three different settings: (i) a sample frequency of $T=128$, and $w_\ell=256$, (ii) $T=w_\ell=256$, and (iii) $T=256$ and $w_\ell=128$. Examples of generated data and respective GT-PCs for cases (i) and (iii) are displayed in Figures \ref{fig:spikes_less} and \ref{fig:spikes_greater} respectively. 

For a downstream comparison of the projections for classification and outlier detection, we generated a second class without spikes. More specifically, we simply used white noise $X^{(t)} = 0.2 \cdot \eps_t$ with distribution $0.2 \cdot \Nc(0, 1)$. For classification, we randomly generated data with a 50\% probability of containing a spike and fitted the projecting models to this mixed data. For outlier detection, we fitted the projecting models only to data with spikes.

The projecting models were trained as before. The ResMSEs are displayed in columns 1--3 of Figure \ref{fig:spikes_results} and Tables \ref{tab:spikes_results_class} and \ref{tab:spikes_results_outlier} in the appendix.   

Based on the projections, we trained downstream MLPs based on mixed data with a spike probability of 50\%. The MLPs were trained for 10 epochs with the Adam optimizer to minimize the categorical crossentropy. The resulting accuracies are displayed on the right of Figure \ref{fig:spikes_results} and Tables \ref{tab:spikes_results_class} and \ref{tab:spikes_results_outlier} in the appendix.  

In all cases, the ResMSEs of PCA and KPCA approximately coincide. Further GT-PCA yields a similar $\resMSE$ as PCA, when the lengths of weights and samples coincide. Again, the ResMSEs are overall lower for the AE and VAE compared to PCA and KPCA. For weights with a different length than the samples, the ResMSEs are approximately the same for the AE, VAE and GT-PCA, yet, for larger weights, the ResMSE of GT-PCA is lower for 1 to 3 components. 

The quality of the projections are reflected in the classification and outlier detection results, as PCA and KPCA have generally the lowest accuracies, the accuracies of AE, VAE and GT-PCA are similar, and the predictions of the GT-PCA based models are perfect even for a single component. Note however, that all PCA-based methods have a slightly worse accuracy for cases (ii) and (iii)

\begin{figure*}[ht]
	\begin{center}
		\centerline{\includegraphics[width=\textwidth]{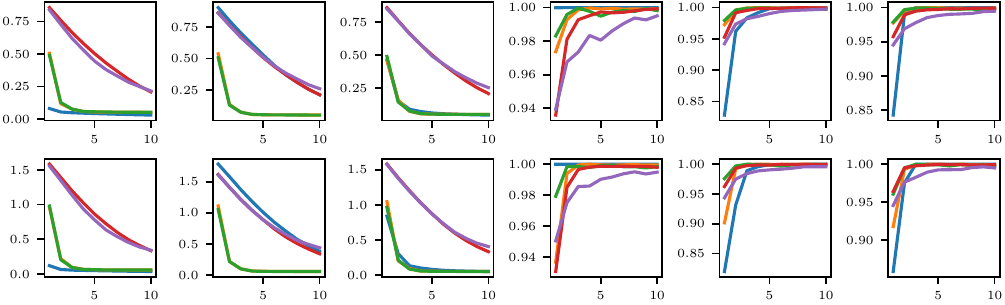}}
		\caption{Results for generated spikes and varying number of components (x-axis) of various models (blue: GT-PCA, red: PCA, purple: KPCA, orange: AE, green: VAE). Top: classification. Bottom: outlier detection. Columns 1 -- 3: Residual MSE (y-axis), respectively settings (i), (ii) and (iii). Columns 4 -- 6: Accuracy (y-axis), respectively settings (i), (ii) and (iii). }
		\label{fig:spikes_results}
	\end{center}
	\vskip -0.3in
\end{figure*}

\subsection{Real Data Experiments} \label{sec:real_data_exp}

For real data experiments, we used the handwriting and MNIST data sets \citep{ramsay2005, lecun1998}, which are common benchmarks in functional data analysis and computer vision, and an EEG data set containing error-related potentials \citep{chavarriaga2010}.

\subsubsection{Projection} \label{sec:real_data_proj}

The handwriting data set contains 20 samples of x-y coordinates, that were recorded along time while writing the letters ``fda''. The samples were recorded over 2.3 seconds, yielding 1401 values per sample (see \citealp{ramsay2005} for a detailed description). To avoid problems with the autoencoders when down- and upsampling, we only used the first 1400 values per sample.

For every iteration of the experiment, we randomly split the data into a training and test set of 16 and 4 samples. Again, we evaluated shift-invariant PCA and used weights of length $1400$. For the samples, we used (i) a window of length $T=700$ at a random position per sample, (ii) the full sample of length $T=1400$.

We trained the (V)AEs for 10 epochs, each consisting of 500 batches of size 32. The GT-PCs were trained sequentially for 10 epochs per component. Each training sample was used 10 000 times during training and  we always used the same window per sample in setting (i). The ResMSEs are displayed Figure \ref{fig:handwriting_results} and Table \ref{tab:handwriting_results} in the appendix.

KPCA yields the highest ResMSE in both settings and does not seem to improve with more components. For setting (i), PCA yields results that are slightly higher compared to AE and VAE for few components, but surpasses the ResMSE of the latter for more components. GT-PCA has overall the lowest ResMSE and is particularly superior for few components. For setting (ii), PCA has the lowest $\resMSE$. Interestingly, the GT-PCs differ from the conventional PCs. This effect is due to a slower training of higher components and it is expected that the results converge for longer trainings. In this case, the results of GT-PCA are similar to those of AE and VAE. 

\begin{figure}[tb]
	\begin{center}
		\centerline{\includegraphics[width=\columnwidth]{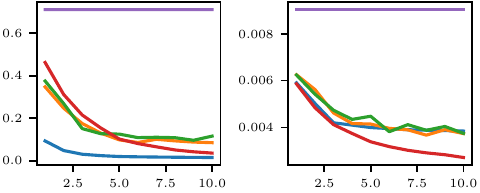}}
		\caption{Residual MSE (y-axis) for handwriting data and varying number of components (x-axis) of various models (blue: GT-PCA, red: PCA, purple: KPCA, orange: AE, green: VAE) Left: Window length $T=700$. Right: Window length $T=1400$.}
		\label{fig:handwriting_results}
	\end{center}
	\vskip -0.3in
\end{figure}

\subsubsection{Classification \& Outlier Detection} \label{sec:real_data_class}

The MNIST data set contains grayscale images of 28x28 pixels with handwritten digits. It is a common benchmark data set in computer vision and is split, by default, into 60 000 training and 10 000 test samples. Additionally to shift-invariant PCA, we tested rotation-invariant components as described in Section \ref{sec:img_data}. With the additional rotation, we have four settings for the MNIST data and always use weights of shape $28\times 28$: (i) we use sub-images of size $16\times 16$ at random positions, (ii) we use the original images, (iii) we embed the original images at random positions in larger images of shape $56\times 56$, (iv) we rotate the original images by a random multiple of $\pi/10$. Examples of the transformed data and respective GT-PCs for cases (iii) and (iv) are displayed in Figure \ref{fig:mnist_shift}.

Again, we first used different projection models and subsequently the same MLP for classification and outlier detection. For classification, we fitted the components to the entire data set while using only the digit ``8'' for the components in case of outlier detection. The entire training data was used for training the subsequent MLPs. 

The number of epochs differed between the settings, but always contained 500 batches of size 32. In cases (i) -- (iii), we trained the projection models for 5 epochs and the subsequent models for 10 epochs. In case (iv), we used 10 epochs for the projection and classification models and 5 epochs for the outlier detection models.

The results for all settings are displayed in Figure \ref{fig:mnist_results} and Tables \ref{tab:mnist_results_class} and \ref{tab:mnist_results_outlier} in the appendix. Further, projections of 1024 randomly rotated digits on their first two rotation-invariant principal components are displayed in Figure \ref{fig:mnist_rot_proj} in the appendix.

As before, the GT-PCs coincide with PCs in setting (ii). In all settings, the ResMSEs are lower for the AE compared to those of the PCA. The ResMSEs of the VAE are much higher than 1 and turn out to be useless. In settings (i), (iii) and (iv), i.\,e., in those where the GT-PCs are not trivial PCs, GT-PCA yields the lowest ResMSEs and is superior to the competing methods. 

These results are also reflected in classification and outlier detection. In particular in cases (i) and (iii), GT-PCA yields the best results.

\begin{figure*}[t]
	\begin{center}
		\centerline{\includegraphics[width=\textwidth]{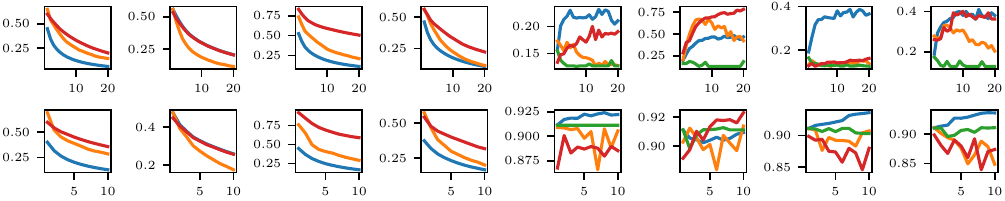}}
		\caption{Results for MNIST data and varying number of components (x-axis) of various models (blue: GT-PCA, red: PCA, orange: AE, green: VAE). Top: classification. Bottom: outlier detection. Columns 1 -- 4: Residual MSE (y-axis), respectively settings (i)-(iv). Columns 5 -- 8: Accuracy (y-axis), respectively settings (i)-(iv). }
		\label{fig:mnist_results}
	\end{center}
	\vskip -0.2in
\end{figure*}

The last data set consists of EEG recordings of 6 subjects with 2 sessions per subject. Per session, 10 blocks of 3 minutes were recorded. Each block consists of approximately 50 trials, where in each trial an error-related potential (ErrP) was provoked with a probability of 20\%. The EEG was recorded with 64 electrodes and a sample frequency of 512\,Hz (see \citealp{chavarriaga2010} for a detailed description). 

As proposed by Chavarriaga \& Millán, we only used electrodes ``Fz'' and ```FCz'' according to the 10-10 system \yrcite{chavarriaga2010}. Further, we used the first recording per subject for training and the second for testing. For comparability, we did not repeat the experiments with random splits, but used the suggested split only. We subtracted the mean per channel to standardize the raw inputs.

ErrPs are expected to last around 0.5\,s, thus, we considered three settings: (i) samples of length 0.5\,s and weights of length 1\,s, (ii) samples and weights of length  0.5\,s and (iii) samples of length 1\,s and weights of length 0.5\,s.

To test GT-PCA for outlier detection, we randomly selected windows between 250\,ms before and 750\,ms after trial onset in settings (i) and (ii), and between 750\,ms before and 1250\,ms after trial onset in setting (iii). In this case, we trained the projection models with data corresponding to ErrPs only. The subsequent MLP was trained with random samples from the entire data set.

When training the projection models for classification, we randomly sampled windows of the given length from the entire data set. Out of the 500 trials per recording, approximately 100 correspond to ErrPs. Each ErrP took about 0.5\,s, which adds up to 50\,s and makes up less than 3\% of the total recording. For the downstream classification, we upsampled the ErrP-part of the data set to have a ratio of approximately 20\% ErrPs and 80\% other EEG signals.

We trained the projection models with 5 epochs of 500 batches of size 32 and the subsequent MLPs with 10 epochs. In previous experiments, KPCA did not yield helpful projections and was excluded from the study. Further, PCA was not used explicitly because conventional principal components are approximated by the GT-PCs in settings (ii). The results are displayed in Figure \ref{fig:errp_results} and Table \ref{tab:errp_results_class} and \ref{tab:errp_results_outlier} of the appendix. 

In all settings, GT-PCA has substantially lower ResMSEs compared to AE and VAE. In particular in case (i), when the weights are larger than the data, GT-PCs seem capable of capturing a major part of the variance. The results of the subsequent classification/outlier detection are less promising, as the trained models were classifying essentially any sample as part of the majority class. 

These results suggest that GT-PCA is able to effectively reduce the dimension of EEG data but the projection based on 10 components is not helpful in the context of ErrPs. 

\begin{figure*}[t]
	\begin{center}
		\centerline{\includegraphics[width=\textwidth]{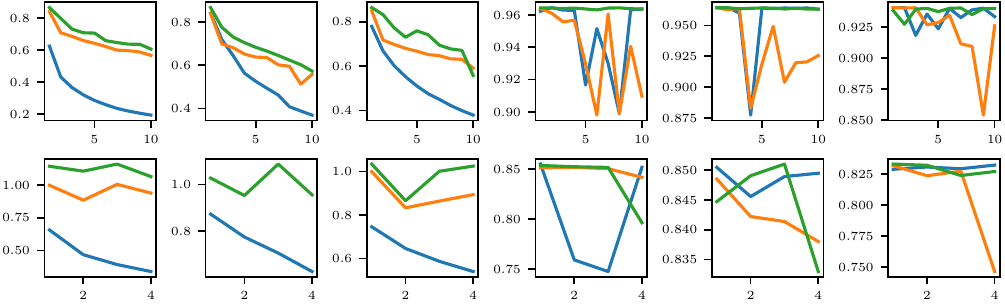}}
		\caption{Results for ErrP data and varying number of components (x-axis) of various models (blue: GT-PCA, orange: AE, green: VAE). Top: classification. Bottom: outlier detection. Columns 1 -- 3: Residual MSE (y-axis), respectively settings (i), (ii) and (iii). Columns 4 -- 6: Accuracy (y-axis), respectively settings (i), (ii) and (iii).}
		\label{fig:errp_results}
	\end{center}
	\vskip -0.2in
\end{figure*}

\section{Conclusion}  \label{sec:con}

In this work, we presented GT-PCA, a generalization of PCA that is invariant w.r.t. problem-dependent transformations. As the components are infeasible to calculate analytically, we proposed a neural network architecture that approximates the components. We conducted experiments with synthesized and real data and could show that GT-PCA outperforms the considered alternatives (PCA, KPCA, AEs and VAEs) for non-trivial families of transformations $\Tc$. As a special case, for $\Tc=\{\ident\}$, we could reproduce conventional PCA. The superior performance of GT-PCA compared to AEs and VAEs is expected to be due to the added inductive bias of the model, which simplifies training. With more components, larger models, more data and longer training, this difference might vanish.

The proposed GT-PCA allows to effectively reduce the dimension of data. Additionally, the components can be easily interpreted, analogously to the components obtained through functional PCA. This makes the proposed GT-PCA specifically helpful in critical applications, where interpretability of predictions is crucial.

When applied to EEG data, GT-PCA seems capable of capturing a major part of the overall signal. However, using two channels and 10 components did not yield promising results in case of ErrP detection. It would be interesting, to see if GT-PCA can detect ErrPs reliably when using more than two EEG channels and possibly more than 10 components. 

The results of this paper suggest that GT-PCA is a relevant area of future research. Besides the detection of ErrPs, GT-PCA might be successfully used in other scenarios for EEG analysis, for instance, the detection of spike waves in the context of epilepsy. Further, GT-PCA might be applied to functional time series, i.\,e., time series where functions are observed at each time point, which includes videos \citep{kidzinski2015}, temperature curves \citep{bucher2020} or intraday prices of stocks \citep{bucher2023}. Finally, other transformations, like monotonic transformations of time, deformations, or a mix of various transformations might be relevant in future applications.

\section*{Acknowledgment}

This work was funded by the Hessian Ministry of Science and the Arts (HMWK) through hessian.AI.

\bibliography{bibliography}
\bibliographystyle{icml2024}

\newpage
\appendix
\onecolumn
\section{Additional Experimental Results}

\begin{table}[H]
	\caption{Residual MSE for oscillating data and varying number of components of various models.}
	\label{tab:oscillations_results}
	\vskip 0.1in
	\begin{center}
		\begin{small}
			\begin{tabular}{l|cccccccccc}
			\toprule
			num. comp. & 1 & 2 & 3 & 4 & 5 & 6 & 7 & 8 & 9 & 10 \\
			\midrule		
			\multicolumn{11}{l}{Weight length $512$}\\		
			GT-PCA & \textbf{0.635} & \textbf{0.464} & \textbf{0.396} & \textbf{0.340} & \textbf{0.295} & \textbf{0.261} & \textbf{0.234} & \textbf{0.213} & \textbf{0.196} & \textbf{0.183}\\
			Autoencoder & 0.898 & 0.729 & 0.565 & 0.482 & 0.454 & 0.440 & 0.432 & 0.428 & 0.422 & 0.416\\
			VAE & 0.886 & 0.704 & 0.564 & 0.489 & 0.452 & 0.441 & 0.433 & 0.428 & 0.422 & 0.416\\
			PCA & 0.926 & 0.854 & 0.779 & 0.705 & 0.632 & 0.560 & 0.504 & 0.448 & 0.425 & 0.404\\
			KPCA & 0.948 & 0.897 & 0.845 & 0.795 & 0.746 & 0.698 & 0.664 & 0.631 & 0.622 & 0.615\\
			\midrule		
			\multicolumn{11}{l}{Weight length $256$}\\		
			GT-PCA & 0.927 & 0.853 & 0.780 & 0.706 & 0.632 & 0.558 & 0.504 & 0.448 & 0.425 & \textbf{0.404}\\
			Autoencoder & 0.899 & 0.716 & \textbf{0.572} & \textbf{0.486} & 0.457 & 0.441 & \textbf{0.433} & \textbf{0.427} & \textbf{0.421} & 0.415\\
			VAE & \textbf{0.892} & \textbf{0.708} & 0.578 & 0.488 & \textbf{0.455} & \textbf{0.440} & \textbf{0.433} & 0.428 & \textbf{0.421} & 0.416\\
			PCA & 0.928 & 0.854 & 0.781 & 0.707 & 0.632 & 0.559 & 0.504 & 0.448 & 0.425 & \textbf{0.404}\\
			KPCA & 0.949 & 0.897 & 0.846 & 0.795 & 0.745 & 0.697 & 0.664 & 0.630 & 0.622 & 0.615\\
			\bottomrule
		\end{tabular}
		\end{small}
	\end{center}
	\vskip -0.1in
\end{table}

\begin{table}[H]
	\caption{Residual MSEs and accuracies of various models for spiking data and varying number of components in the classification setting.}
	\label{tab:spikes_results_class}
	\vskip 0.1in
	\begin{center}
		\begin{small}
			\begin{tabular}{l|cccccccccc}
				\toprule
				num. comp. & 1 & 2 & 3 & 4 & 5 & 6 & 7 & 8 & 9 & 10 \\
				\midrule		
				\multicolumn{11}{l}{ResMSEs -- (i) $T=128, w_\ell= 256$}\\		
				GT-PCA & \textbf{0.081} & \textbf{0.054} & \textbf{0.048} & \textbf{0.045} & \textbf{0.042} & \textbf{0.040} & \textbf{0.037} & \textbf{0.035} & \textbf{0.032} & \textbf{0.030}\\
				Autoencoder & 0.506 & 0.118 & 0.076 & 0.058 & 0.055 & 0.053 & 0.054 & 0.053 & 0.052 & 0.052\\
				VAE & 0.491 & 0.128 & 0.077 & 0.058 & 0.056 & 0.055 & 0.053 & 0.052 & 0.053 & 0.052\\
				PCA & 0.857 & 0.760 & 0.667 & 0.582 & 0.504 & 0.433 & 0.369 & 0.307 & 0.253 & 0.209\\
				KPCA & 0.840 & 0.729 & 0.621 & 0.532 & 0.446 & 0.381 & 0.332 & 0.284 & 0.249 & 0.216\\
				\midrule		
				\multicolumn{11}{l}{ResMSEs -- (ii) $T=256=w_\ell= 256$}\\		
				GT-PCA & 0.903 & 0.799 & 0.703 & 0.615 & 0.527 & 0.445 & 0.371 & 0.311 & 0.257 & 0.213\\
				Autoencoder & 0.538 & 0.137 & \textbf{0.075} & \textbf{0.057} & \textbf{0.055} & 0.054 & \textbf{0.053} & \textbf{0.053} & 0.053 & 0.053\\
				VAE & \textbf{0.506} & \textbf{0.131} & \textbf{0.075} & 0.058 & \textbf{0.055} & \textbf{0.053} & \textbf{0.053} & \textbf{0.053} & \textbf{0.052} & \textbf{0.052}\\
				PCA & 0.862 & 0.765 & 0.675 & 0.588 & 0.507 & 0.435 & 0.372 & 0.312 & 0.258 & 0.214\\
				KPCA & 0.860 & 0.763 & 0.668 & 0.588 & 0.511 & 0.441 & 0.384 & 0.338 & 0.293 & 0.261\\
				\midrule		
				\multicolumn{11}{l}{ResMSEs -- (iii) $T=256, w_\ell= 128$}\\		
				GT-PCA & 0.464 & 0.154 & 0.091 & 0.073 & 0.063 & 0.057 & 0.053 & \textbf{0.050} & \textbf{0.048} & \textbf{0.047}\\
				Autoencoder & \textbf{0.458} & 0.160 & \textbf{0.075} & \textbf{0.056} & \textbf{0.054} & \textbf{0.053} & 0.053 & 0.052 & 0.052 & 0.052\\
				VAE & 0.490 & \textbf{0.145} & 0.078 & 0.057 & \textbf{0.054} & \textbf{0.053} & \textbf{0.052} & 0.052 & 0.052 & 0.052\\
				PCA & 0.859 & 0.760 & 0.664 & 0.579 & 0.501 & 0.430 & 0.365 & 0.305 & 0.253 & 0.209\\
				KPCA & 0.850 & 0.757 & 0.661 & 0.578 & 0.493 & 0.434 & 0.376 & 0.329 & 0.287 & 0.254\\
				\midrule		
				\multicolumn{11}{l}{Accuracies -- (i) $T=128, w_\ell= 256$}\\		
				GT-PCA & \textbf{100.0} & \textbf{100.0} & \textbf{100.0} & \textbf{100.0} & \textbf{100.0} & \textbf{100.0} & \textbf{100.0} & \textbf{100.0} & \textbf{100.0} & \textbf{100.0}\\
				Autoencoder & 97.4 & 99.3 & 99.9 & \textbf{100.0} & 99.9 & \textbf{100.0} & 99.9 & 99.9 & 99.9 & 99.9\\
				VAE & 98.3 & 99.6 & \textbf{100.0} & 99.8 & 99.5 & 99.7 & 99.8 & 99.8 & 99.9 & 99.8\\
				PCA & 93.6 & 98.1 & 99.3 & 99.5 & 99.7 & 99.7 & 99.7 & 99.8 & 99.9 & 99.9\\
				KPCA & 93.9 & 96.8 & 97.4 & 98.3 & 98.1 & 98.6 & 99.0 & 99.4 & 99.3 & 99.5\\
				\midrule		
				\multicolumn{11}{l}{Accuracies -- (ii) $T=w_\ell= 256$}\\		
				GT-PCA & 82.8 & 96.2 & 98.4 & 99.2 & 99.8 & \textbf{99.9} & \textbf{100.0} & \textbf{100.0} & \textbf{100.0} & \textbf{100.0}\\
				Autoencoder & 97.2 & 99.5 & \textbf{99.9} & \textbf{100.0} & \textbf{99.9} & \textbf{99.9} & 99.9 & 99.9 & 99.9 & 99.9\\
				VAE & \textbf{97.9} & \textbf{99.6} & \textbf{99.9} & 99.9 & \textbf{99.9} & 99.8 & 99.8 & 99.9 & 99.8 & 99.7\\
				PCA & 95.2 & 99.2 & 99.5 & 99.8 & 99.8 & \textbf{99.9} & 99.9 & \textbf{100.0} & 99.9 & 99.9\\
				KPCA & 94.2 & 97.4 & 98.2 & 98.6 & 99.0 & 99.4 & 99.5 & 99.6 & 99.7 & 99.7\\
				\midrule		
				\multicolumn{11}{l}{Accuracies -- (iii) $T=256, w_\ell= 256$}\\		
				GT-PCA & 84.3 & 99.0 & \textbf{100.0} & \textbf{100.0} & \textbf{100.0} & \textbf{100.0} & \textbf{100.0} & \textbf{100.0} & \textbf{100.0} & \textbf{100.0}\\
				Autoencoder & 97.8 & 99.5 & 99.9 & \textbf{100.0} & \textbf{100.0} & 99.9 & 99.9 & 99.9 & 99.9 & 99.9\\
				VAE & \textbf{97.9} & \textbf{99.7} & 99.9 & 99.9 & 99.9 & 99.8 & 99.8 & 99.8 & 99.5 & 99.8\\
				PCA & 95.7 & 98.9 & 99.5 & 99.7 & 99.7 & 99.8 & 99.8 & 99.9 & 99.9 & 99.9\\
				KPCA & 94.5 & 96.9 & 97.8 & 98.5 & 98.8 & 99.0 & 99.1 & 99.1 & 99.4 & 99.5\\
				\bottomrule
				\end{tabular}
			\end{small}
		\end{center}
	\vskip -0.1in
\end{table}
				
\begin{table}[H]
	\caption{Residual MSEs and accuracies of various models for spiking data and varying number of components in the outlier detection setting.}
	\label{tab:spikes_results_outlier}
	\vskip 0.1in
	\begin{center}
		\begin{small}
			\begin{tabular}{l|cccccccccc}
				\toprule
				num. comp. & 1 & 2 & 3 & 4 & 5 & 6 & 7 & 8 & 9 & 10 \\
				\midrule		
				\multicolumn{11}{l}{ResMSEs -- (i) $T=128, w_\ell= 256$}\\		
				GT-PCA & \textbf{0.118} & \textbf{0.063} & \textbf{0.054} & \textbf{0.050} & \textbf{0.047} & \textbf{0.044} & \textbf{0.041} & \textbf{0.039} & \textbf{0.036} & \textbf{0.034}\\
				Autoencoder & 0.960 & 0.221 & 0.094 & 0.066 & 0.061 & 0.057 & 0.057 & 0.057 & 0.056 & 0.056\\
				VAE & 0.980 & 0.204 & 0.093 & 0.064 & 0.061 & 0.057 & 0.057 & 0.056 & 0.056 & 0.055\\
				PCA & 1.585 & 1.393 & 1.199 & 1.026 & 0.863 & 0.726 & 0.601 & 0.494 & 0.409 & 0.335\\
				KPCA & 1.558 & 1.349 & 1.137 & 0.937 & 0.780 & 0.636 & 0.538 & 0.455 & 0.387 & 0.342\\
				\midrule		
				\multicolumn{11}{l}{ResMSEs -- (ii) $T=w_\ell= 256$}\\		
				GT-PCA & 1.783 & 1.581 & 1.373 & 1.182 & 1.005 & 0.844 & 0.704 & 0.573 & 0.469 & 0.379\\
				Autoencoder & 1.110 & \textbf{0.215} & 0.099 & 0.064 & 0.059 & 0.058 & 0.057 & 0.056 & 0.056 & 0.056\\
				VAE & \textbf{1.051} & 0.216 & \textbf{0.098} & \textbf{0.062} & \textbf{0.058} & \textbf{0.056} & \textbf{0.055} & \textbf{0.055} & \textbf{0.055} & \textbf{0.055}\\
				PCA & 1.616 & 1.410 & 1.216 & 1.041 & 0.885 & 0.737 & 0.610 & 0.506 & 0.417 & 0.340\\
				KPCA & 1.607 & 1.408 & 1.210 & 1.036 & 0.885 & 0.757 & 0.651 & 0.556 & 0.492 & 0.437\\
				\midrule		
				\multicolumn{11}{l}{ResMSEs -- (iii) $T=256, w_\ell= 128$}\\		
				GT-PCA & \textbf{0.841} & 0.307 & 0.136 & 0.104 & 0.084 & 0.072 & 0.064 & 0.060 & 0.056 & \textbf{0.054}\\
				Autoencoder & 1.045 & \textbf{0.205} & 0.097 & \textbf{0.060} & 0.058 & \textbf{0.056} & \textbf{0.055} & 0.056 & 0.055 & 0.055\\
				VAE & 0.964 & 0.214 & \textbf{0.087} & 0.061 & \textbf{0.057} & \textbf{0.056} & \textbf{0.055} & \textbf{0.054} & \textbf{0.054} & \textbf{0.054}\\
				PCA & 1.580 & 1.385 & 1.203 & 1.029 & 0.874 & 0.727 & 0.602 & 0.495 & 0.408 & 0.335\\
				KPCA & 1.574 & 1.375 & 1.194 & 1.025 & 0.870 & 0.735 & 0.612 & 0.526 & 0.455 & 0.408\\
				\midrule		
				\multicolumn{11}{l}{Accuracies -- (i) $T=128, w_\ell= 256$}\\		
				GT-PCA & \textbf{100.0} & \textbf{100.0} & \textbf{100.0} & \textbf{100.0} & \textbf{100.0} & \textbf{100.0} & \textbf{100.0} & \textbf{100.0} & \textbf{100.0} & \textbf{100.0}\\
				Autoencoder & 93.7 & 99.4 & 99.9 & \textbf{100.0} & \textbf{100.0} & 99.9 & \textbf{100.0} & \textbf{100.0} & 99.9 & 99.9\\
				VAE & 97.9 & 99.8 & 99.8 & 99.8 & 99.9 & 99.9 & 99.8 & 99.8 & 99.8 & 99.8\\
				PCA & 93.1 & 98.5 & 99.7 & 99.8 & 99.9 & 99.9 & 99.9 & 99.9 & 99.9 & 99.8\\
				KPCA & 95.0 & 97.5 & 98.6 & 98.6 & 99.0 & 99.2 & 99.4 & 99.5 & 99.4 & 99.5\\
				\midrule		
				\multicolumn{11}{l}{Accuracies -- (ii) $T= w_\ell= 256$}\\		
				GT-PCA & 82.0 & 93.2 & 98.9 & 99.6 & \textbf{99.9} & \textbf{99.9} & \textbf{99.9} & 99.9 & \textbf{100.0} & \textbf{100.0}\\
				Autoencoder & 90.1 & 99.2 & 99.9 & \textbf{100.0} & \textbf{99.9} & \textbf{99.9} & \textbf{99.9} & \textbf{100.0} & 99.9 & 99.9\\
				VAE & \textbf{97.5} & \textbf{99.7} & \textbf{100.0} & 99.9 & 99.8 & 99.6 & 99.8 & 99.7 & 99.9 & 99.8\\
				PCA & 96.2 & 99.3 & 99.7 & 99.8 & 99.8 & \textbf{99.9} & \textbf{99.9} & 99.9 & 99.9 & 99.9\\
				KPCA & 94.3 & 97.5 & 98.4 & 98.8 & 99.0 & 99.1 & 99.3 & 99.5 & 99.5 & 99.5\\
				\midrule		
				\multicolumn{11}{l}{Accuracies -- (iii) $T=256, w_\ell= 128$}\\		
				GT-PCA & 85.9 & 97.5 & \textbf{100.0} & \textbf{100.0} & \textbf{100.0} & \textbf{100.0} & \textbf{100.0} & \textbf{100.0} & \textbf{100.0} & \textbf{100.0}\\
				Autoencoder & 91.8 & 99.3 & 99.9 & \textbf{100.0} & \textbf{100.0} & 99.9 & \textbf{100.0} & 99.9 & 99.9 & 99.9\\
				VAE & 96.0 & \textbf{99.5} & \textbf{100.0} & 99.9 & 99.9 & 99.8 & 99.9 & 99.8 & 99.9 & 99.7\\
				PCA & \textbf{96.4} & \textbf{99.5} & 99.8 & 99.9 & 99.9 & 99.9 & 99.9 & 99.9 & 99.9 & 99.9\\
				KPCA & 94.6 & 97.6 & 98.3 & 98.9 & 99.2 & 99.3 & 99.3 & 99.5 & 99.6 & 99.5\\
				\bottomrule
			\end{tabular}
		\end{small}
	\end{center}
	\vskip -0.1in
\end{table}

\begin{table}[H]
	\caption{Residual MSEs for handwriting data and varying number of components of various models.}
	\label{tab:handwriting_results}
	\vskip 0.1in
	\begin{center}
		\begin{small}
			\begin{tabular}{l|cccccccccc}
				\toprule
				num. comp. & 1 & 2 & 3 & 4 & 5 & 6 & 7 & 8 & 9 & 10 \\
				\midrule		
				\multicolumn{11}{l}{Window length $T=700$}\\		
				GT-PCA & \textbf{0.09289} & \textbf{0.04768} & \textbf{0.03027} & \textbf{0.02373} & \textbf{0.01955} & \textbf{0.01809} & \textbf{0.01703} & \textbf{0.01623} & \textbf{0.01544} & \textbf{0.01495}\\
				Autoencoder & 0.34750 & 0.24722 & 0.17495 & 0.12985 & 0.09788 & 0.08576 & 0.10070 & 0.09320 & 0.08756 & 0.08471\\
				VAE & 0.37506 & 0.26881 & 0.15104 & 0.12698 & 0.12507 & 0.10855 & 0.11028 & 0.10824 & 0.09604 & 0.11565\\
				PCA & 0.46235 & 0.31163 & 0.21501 & 0.15390 & 0.10174 & 0.08047 & 0.06520 & 0.05050 & 0.04179 & 0.03567\\
				KPCA & 0.71136 & 0.71136 & 0.71136 & 0.71136 & 0.71136 & 0.71136 & 0.71136 & 0.71136 & 0.71136 & 0.71136\\
				\midrule		
				\multicolumn{11}{l}{Window length $T=1400$}\\		
				GT-PCA & 0.00591 & 0.00502 & 0.00420 & 0.00410 & 0.00400 & 0.00395 & 0.00390 & 0.00388 & 0.00387 & 0.00384\\
				Autoencoder & 0.00625 & 0.00562 & 0.00462 & 0.00417 & 0.00415 & 0.00396 & 0.00390 & 0.00367 & 0.00390 & 0.00374\\
				VAE & 0.00624 & 0.00542 & 0.00473 & 0.00434 & 0.00448 & 0.00382 & 0.00412 & 0.00388 & 0.00404 & 0.00375\\
				PCA & \textbf{0.00586} & \textbf{0.00484} & \textbf{0.00412} & \textbf{0.00374} & \textbf{0.00339} & \textbf{0.00318} & \textbf{0.00302} & \textbf{0.00291} & \textbf{0.00283} & \textbf{0.00271}\\
				KPCA & 0.00904 & 0.00904 & 0.00904 & 0.00904 & 0.00904 & 0.00904 & 0.00904 & 0.00904 & 0.00904 & 0.00904\\
				\bottomrule
			\end{tabular}
		\end{small}
	\end{center}
	\vskip -0.1in
\end{table}

\begin{table}[H]
	\caption{Residual MSEs and accuracies of various models for ErrP data and varying number of components in the classification setting.}
	\label{tab:errp_results_class}
	\vskip 0.1in
	\begin{center}
		\begin{small}
			\begin{tabular}{l|cccccccccc}
				\toprule
				num. comp. & 1 & 2 & 3 & 4 & 5 & 6 & 7 & 8 & 9 & 10 \\
				\midrule		
				\multicolumn{11}{l}{ResMSEs -- (i) sample length: 0.5\,s, weight length: 1\,s}\\		
				GT-PCA      & \textbf{0.624} & \textbf{0.431} & \textbf{0.364} & \textbf{0.319} & \textbf{0.284} & \textbf{0.257} & \textbf{0.234} & \textbf{0.217} & \textbf{0.204} & \textbf{0.192}\\
				Autoencoder & 0.841 & 0.707 & 0.685 & 0.659 & 0.641 & 0.621 & 0.599 & 0.596 & 0.587 & 0.566\\
				VAE         & 0.865 & 0.798 & 0.731 & 0.708 & 0.705 & 0.658 & 0.646 & 0.636 & 0.636 & 0.605\\
				\multicolumn{11}{l}{ResMSEs -- (ii) sample and weight length: 0.5\,s}\\		
				GT-PCA      & 0.841 & 0.719 & \textbf{0.645} & \textbf{0.564} & \textbf{0.526} & \textbf{0.494} & \textbf{0.463} & \textbf{0.408} & \textbf{0.389} & \textbf{0.370}\\
				Autoencoder & \textbf{0.839} & \textbf{0.698} & 0.681 & 0.652 & 0.638 & 0.635 & 0.602 & 0.595 & 0.513 & 0.558\\
				VAE         & 0.866 & 0.775 & 0.730 & 0.705 & 0.683 & 0.665 & 0.645 & 0.624 & 0.603 & 0.573\\
				\multicolumn{11}{l}{ResMSEs -- (iii) sample length: 1\,s, weight length: 0.5\,s}\\		
				GT-PCA      & \textbf{0.779} & \textbf{0.668} & \textbf{0.601} & \textbf{0.552} & \textbf{0.511} & \textbf{0.476} & \textbf{0.450} & \textbf{0.423} & \textbf{0.399} & \textbf{0.379}\\
				Autoencoder & 0.849 & 0.717 & 0.697 & 0.680 & 0.667 & 0.652 & 0.647 & 0.633 & 0.631 & 0.591\\
				VAE         & 0.864 & 0.832 & 0.772 & 0.730 & 0.759 & 0.741 & 0.694 & 0.678 & 0.671 & 0.558\\
				\multicolumn{11}{l}{Accuracies -- (i) sample length: 0.5\,s, weight length: 1\,s}\\		
				GT-PCA      & 96.2 & \textbf{96.4} & 96.3 & 96.3 & 91.7 & 95.1 & 93.0 & 89.9 & 96.3 & \textbf{96.4}\\
				Autoencoder & \textbf{96.4} & 96.1 & 95.6 & 95.7 & 93.0 & 89.8 & 96.1 & 89.9 & 94.1 & 91.0\\
				VAE         & \textbf{96.4} & \textbf{96.4} & \textbf{96.4} & \textbf{96.4} & \textbf{96.4} & \textbf{96.3} & \textbf{96.4} & \textbf{96.4} & \textbf{96.4} & \textbf{96.4}\\
				\multicolumn{11}{l}{Accuracies -- (ii) sample and weight length: 0.5\,s}\\		
				GT-PCA      & \textbf{96.4} & \textbf{96.4} & 96.0 & 87.7 & \textbf{96.4} & \textbf{96.4} & \textbf{96.4} & \textbf{96.4} & \textbf{96.4} & \textbf{96.3}\\
				Autoencoder & \textbf{96.4} & 96.3 & \textbf{96.4} & 88.3 & 91.9 & 94.9 & 90.4 & 92.0 & 92.0 & 92.6\\
				VAE         & \textbf{96.4} & \textbf{96.4} & 96.3 & \textbf{96.4} & \textbf{96.4} & \textbf{96.4} & 96.3 & \textbf{96.4} & 96.3 & \textbf{96.3}\\
				\multicolumn{11}{l}{Accuracies -- (iii) sample length: 1\,s, weight length: 0.5\,s}\\		
				GT-PCA      & \textbf{94.0} & \textbf{94.0} & 91.8 & 93.5 & 92.3 & \textbf{94.0} & 93.2 & \textbf{93.9} & \textbf{94.0} & 93.3\\
				Autoencoder & \textbf{94.0} & \textbf{94.0} & \textbf{94.0} & 92.7 & 92.8 & 93.4 & 91.1 & 90.9 & 85.4 & 92.6\\
				VAE         & 93.8 & 92.7 & 93.9 & \textbf{94.0} & \textbf{93.7} & \textbf{94.0} & \textbf{94.0} & 93.5 & \textbf{94.0} & \textbf{94.0}\\
				\bottomrule
			\end{tabular}
		\end{small}
	\end{center}
	\vskip -0.1in
\end{table}

\begin{table}[H]
	\caption{Residual MSEs and accuracies of various models for ErrP data and varying number of components in the outlier detection setting.}
	\label{tab:errp_results_outlier}
	\vskip 0.1in
	\begin{center}
		\begin{small}
			\begin{tabular}{l|cccc}
				\toprule
				num. comp. & 1 & 2 & 3 & 4 \\
				\midrule		
				\multicolumn{5}{l}{ResMSEs -- (i) sample length: 0.5\,s, weight length: 1\,s}\\		
				GT-PCA & \textbf{0.657} & \textbf{0.469} & \textbf{0.394} & \textbf{0.341}\\
				Autoencoder & 0.996 & 0.883 & 1.003 & 0.938\\
				VAE & 1.141 & 1.104 & 1.158 & 1.062\\
				\midrule		
				\multicolumn{5}{l}{ResMSEs -- (ii) sample and weight length: 0.5\,s}\\		
				GT-PCA & \textbf{0.872} & \textbf{0.774} & \textbf{0.704} & \textbf{0.625}\\
				Autoencoder & 17.514 & 0.910 & 0.942 & 0.930\\
				VAE & 1.027 & 0.952 & 1.088 & 0.955\\
				\midrule		
				\multicolumn{5}{l}{ResMSEs -- (iii) sample length: 1\,s, weight length: 0.5\,s}\\		
				GT-PCA & \textbf{0.746} & \textbf{0.647} & \textbf{0.588} & \textbf{0.541}\\
				Autoencoder & 0.999 & 0.833 & 0.864 & 0.894\\
				VAE & 1.034 & 0.866 & 1.001 & 1.025\\
				\midrule		
				\multicolumn{5}{l}{Accuracies -- (i) sample length: 0.5\,s, weight length: 1\,s}\\		
				GT-PCA & \textbf{85.5} & 75.9 & 74.8 & \textbf{85.2}\\
				Autoencoder & 85.1 & 85.2 & 85.1 & 84.2\\
				VAE & 85.4 & \textbf{85.3} & \textbf{85.2} & 79.7\\
				\midrule		
				\multicolumn{5}{l}{Accuracies -- (ii) sample and weight length: 0.5\,s}\\		
				GT-PCA & \textbf{85.0} & 84.6 & 84.9 & \textbf{84.9}\\
				Autoencoder & 84.8 & 84.2 & 84.1 & 83.8\\
				VAE & 84.5 & \textbf{84.9} & \textbf{85.1} & 83.3\\
				\midrule		
				\multicolumn{5}{l}{Accuracies -- (iii) sample length: 1\,s, weight length: 0.5\,s}\\		
				GT-PCA & 82.9 & 83.0 & \textbf{82.9} & \textbf{83.2}\\
				Autoencoder & 83.2 & 82.4 & 82.7 & 74.6\\
				VAE & \textbf{83.3} & \textbf{83.2} & 82.4 & 82.7\\
				\bottomrule
			\end{tabular}
		\end{small}
	\end{center}
	\vskip -0.1in
\end{table}

\setlength\tabcolsep{2 pt}
\begin{table}[H]
	\caption{Residual MSEs and accuracies of various models for MNIST data and varying number of components in the classification setting. ``--'' indicates that the gradients exploded for all repetitions.}
	\label{tab:mnist_results_class}
	\vskip 0.1in
	\begin{center}
		\begin{small}
			\rotatebox{90}{
				\begin{tabular}{l|cccccccccccccccccccc}
					\toprule
					num. comp. & 1 & 2 & 3 & 4 & 5 & 6 & 7 & 8 & 9 & 10 & 11 & 12 & 13 & 14 & 15 & 16 & 17 & 18 & 19 & 20 \\
					\midrule		
					\multicolumn{21}{l}{ResMSEs -- (i) samples of size $16\times 16$}\\		
					GT-PCA      & \textbf{0.449} & \textbf{0.343} & \textbf{0.280} & \textbf{0.239} & \textbf{0.208} & \textbf{0.184} & \textbf{0.165} & \textbf{0.149} & \textbf{0.136} & \textbf{0.125} & \textbf{0.116} & \textbf{0.107} & \textbf{0.100} & \textbf{0.093} & \textbf{0.088} & \textbf{0.082} & \textbf{0.078} & \textbf{0.073} & \textbf{0.069} & \textbf{0.066}\\
					Autoencoder & 0.646 & 0.537 & 0.487 & 0.436 & 0.391 & 0.350 & 0.316 & 0.293 & 0.272 & 0.253 & 0.232 & 0.221 & 0.207 & 0.195 & 0.185 & 0.171 & 0.163 & 0.157 & 0.152 & 0.147\\
					VAE         & 0.694 & 364.278 & 1.022 & -- & -- & -- & 4992.540 & -- & -- & 1.001 & 1.617 & -- & 1.001 & -- & -- & -- & -- & 196.473 & -- & -- \\
					PCA         & 0.597 & 0.552 & 0.514 & 0.478 & 0.444 & 0.418 & 0.393 & 0.371 & 0.351 & 0.332 & 0.315 & 0.299 & 0.284 & 0.269 & 0.256 & 0.244 & 0.233 & 0.223 & 0.213 & 0.203\\
					\midrule		
					\multicolumn{21}{l}{ResMSEs -- (ii) samples of size $28\times 28$}\\		
					GT-PCA      & 0.557 & 0.508 & 0.464 & 0.427 & 0.395 & 0.370 & 0.349 & 0.331 & 0.315 & 0.301 & 0.288 & 0.276 & 0.266 & 0.256 & 0.246 & 0.237 & 0.229 & 0.222 & 0.215 & 0.208\\
					Autoencoder & 0.557 & \textbf{0.463} & \textbf{0.392} & \textbf{0.330} & \textbf{0.291} & \textbf{0.259} & \textbf{0.235} & \textbf{0.217} & \textbf{0.202} & \textbf{0.188} & \textbf{0.178} & \textbf{0.167} & \textbf{0.159} & \textbf{0.150} & \textbf{0.142} & \textbf{0.137} & \textbf{0.133} & \textbf{0.127} & \textbf{0.121} & \textbf{0.118}\\
					VAE         & 0.764 & 0.743 & 0.654 & 0.969 & 0.751 & -- & 0.896 & 1.284 & -- & -- & -- & -- & -- & -- & -- & 1.000 & -- & -- & 1.001 & 649.490\\
					PCA         & \textbf{0.533} & 0.489 & 0.453 & 0.420 & 0.391 & 0.365 & 0.346 & 0.329 & 0.312 & 0.299 & 0.287 & 0.274 & 0.264 & 0.254 & 0.245 & 0.236 & 0.228 & 0.220 & 0.213 & 0.207\\
					\midrule		
					\multicolumn{21}{l}{ResMSEs -- (iii) samples of size $56\times 56$}\\		
					GT-PCA      & \textbf{0.525} & \textbf{0.427} & \textbf{0.357} & \textbf{0.314} & \textbf{0.282} & \textbf{0.255} & \textbf{0.234} & \textbf{0.216} & \textbf{0.200} & \textbf{0.187} & \textbf{0.175} & \textbf{0.165} & \textbf{0.156} & \textbf{0.148} & \textbf{0.141} & \textbf{0.134} & \textbf{0.128} & \textbf{0.123} & \textbf{0.118} & \textbf{0.113}\\
					Autoencoder & 0.736 & 0.640 & 0.557 & 0.518 & 0.482 & 0.448 & 0.414 & 0.387 & 0.358 & 0.330 & 0.312 & 0.298 & 0.283 & 0.274 & 0.260 & 0.252 & 0.241 & 0.229 & 0.220 & 0.211\\
					VAE         & 0.917 & 0.818 & 0.927 & 0.988 & 0.902 & 0.774 & 0.870 & 0.862 & 53.232 & 0.817 & 1.053 & 1.889 & 1.212 & 0.899 & 1.064 & 0.768 & 1.007 & 3.387 & 1.284 & 1.392\\
					PCA         & 0.831 & 0.772 & 0.726 & 0.688 & 0.665 & 0.643 & 0.626 & 0.613 & 0.600 & 0.590 & 0.581 & 0.571 & 0.562 & 0.552 & 0.544 & 0.536 & 0.528 & 0.521 & 0.513 & 0.506\\
					\midrule		
					\multicolumn{21}{l}{ResMSEs -- (iv) randomly rotated samples}\\		
					GT-PCA      & \textbf{0.460} & \textbf{0.376} & \textbf{0.320} & \textbf{0.282} & \textbf{0.255} & \textbf{0.233} & \textbf{0.216} & \textbf{0.201} & \textbf{0.188} & \textbf{0.176} & \textbf{0.166} & \textbf{0.157} & \textbf{0.149} & \textbf{0.141} & \textbf{0.134} & \textbf{0.128} & \textbf{0.122} & \textbf{0.117} & \textbf{0.112} & \textbf{0.107}\\
					Autoencoder & 0.561 & 0.512 & 0.448 & 0.401 & 0.361 & 0.319 & 0.291 & 0.265 & 0.242 & 0.224 & 0.207 & 0.191 & 0.177 & 0.163 & 0.154 & 0.144 & 0.138 & 0.129 & 0.121 & 0.115\\
					VAE         & 0.721 & 0.690 & 0.701 & -- & 595.688 & -- & 1.000 & 1.000 & 579.101 & 0.481 & -- & -- & -- & -- & -- & 3398.896 & 1.000 & 1.620 & -- & -- \\
					PCA         & 0.560 & 0.518 & 0.476 & 0.448 & 0.427 & 0.406 & 0.386 & 0.366 & 0.347 & 0.332 & 0.316 & 0.301 & 0.285 & 0.274 & 0.263 & 0.254 & 0.245 & 0.237 & 0.230 & 0.222\\
					\midrule		
					\multicolumn{21}{l}{Accuracies -- (i) samples of size $16\times 16$}\\		
					GT-PCA      & 15.7 & \textbf{20.1} & \textbf{21.3} & \textbf{21.9} & \textbf{23.0} & \textbf{21.7} & \textbf{21.5} & \textbf{21.6} & \textbf{21.6} & \textbf{21.6} & \textbf{22.1} & \textbf{21.3} & \textbf{21.8} & \textbf{23.1} & \textbf{22.2} & \textbf{23.0} & \textbf{22.8} & \textbf{21.1} & \textbf{20.4} & \textbf{21.0}\\
					Autoencoder & \textbf{17.3} & 16.5 & 15.5 & 17.1 & 16.0 & 16.7 & 15.2 & 14.6 & 14.3 & 14.2 & 14.0 & 13.7 & 12.7 & 13.2 & 12.8 & 13.5 & 12.9 & 12.9 & 12.9 & 12.8\\
					VAE         & 15.5 & 13.8 & 13.2 & 12.7 & 12.7 & 12.7 & 12.6 & 12.7 & 12.7 & 12.7 & 13.1 & 12.7 & 12.7 & 12.7 & 12.7 & 12.7 & 12.7 & 13.7 & 12.7 & 12.7\\
					PCA         & 13.4 & 14.8 & 14.9 & 16.0 & 16.5 & 16.3 & 16.8 & 18.0 & 17.5 & 17.1 & 18.1 & 20.0 & 17.5 & 19.3 & 18.0 & 18.7 & 18.6 & 18.8 & 18.5 & 19.0\\
					\midrule		
					\multicolumn{21}{l}{Accuracies -- (ii) samples of size $28\times 28$}\\		
					GT-PCA      & 19.8 & 26.8 & 33.2 & 35.8 & 39.5 & 42.4 & 42.6 & 44.9 & 45.1 & 46.6 & 46.6 & 44.1 & 45.1 & 47.5 & 45.7 & 46.2 & 48.3 & 45.2 & 43.6 & 46.6\\
					Autoencoder & 20.4 & 36.9 & \textbf{49.5} & \textbf{59.5} & \textbf{66.2} & \textbf{67.2} & \textbf{66.1} & 61.0 & 66.9 & 63.0 & 55.3 & 59.5 & 57.0 & 58.6 & 47.5 & 52.7 & 42.5 & 46.6 & 47.1 & 42.3\\
					VAE         & 16.0 & 16.1 & 18.4 & 17.6 & 15.4 & 12.7 & 14.1 & 17.4 & 12.7 & 12.7 & 12.7 & 12.7 & 12.7 & 12.7 & 12.7 & 12.7 & 12.7 & 12.7 & 12.7 & 18.2\\
					PCA         & \textbf{28.4} & \textbf{40.1} & 42.2 & 50.8 & 58.7 & 62.8 & 66.0 & \textbf{68.4} & \textbf{68.4} & \textbf{69.6} & \textbf{69.9} & \textbf{72.3} & \textbf{73.9} & \textbf{74.5} & \textbf{72.7} & \textbf{73.5} & \textbf{73.8} & \textbf{77.5} & \textbf{76.4} & \textbf{78.4}\\
					\midrule		
					\multicolumn{21}{l}{Accuracies -- (iii) samples of size $56\times 56$}\\		
					GT-PCA      & \textbf{19.1} & \textbf{25.7} & \textbf{31.6} & \textbf{33.9} & \textbf{34.0} & \textbf{35.2} & \textbf{34.9} & \textbf{34.7} & \textbf{34.5} & \textbf{38.0} & \textbf{35.7} & \textbf{37.3} & \textbf{37.4} & \textbf{38.7} & \textbf{35.2} & \textbf{37.4} & \textbf{38.6} & \textbf{38.0} & \textbf{35.9} & \textbf{36.6}\\
					Autoencoder & 13.3 & 15.2 & 14.0 & 13.2 & 13.2 & 13.0 & 13.1 & 12.5 & 13.8 & 14.1 & 13.5 & 14.3 & 13.8 & 14.0 & 14.1 & 13.7 & 15.1 & 15.0 & 15.0 & 13.7\\
					VAE         & 16.4 & 14.1 & 13.8 & 13.5 & 12.9 & 12.8 & 12.9 & 12.8 & 13.0 & 13.1 & 12.7 & 12.7 & 13.3 & 12.6 & 12.7 & 12.7 & 13.1 & 12.8 & 12.7 & 12.6\\
					PCA         & 12.8 & 13.8 & 12.5 & 13.1 & 13.8 & 13.7 & 14.5 & 14.4 & 14.6 & 14.1 & 14.2 & 14.2 & 14.6 & 15.3 & 15.1 & 15.2 & 15.4 & 15.1 & 15.8 & 16.2\\
					\midrule		
					\multicolumn{21}{l}{Accuracies -- (iv) randomly rotated samples}\\		
					GT-PCA      & 18.8 & \textbf{29.3} & \textbf{32.3} & \textbf{34.6} & \textbf{35.2} & \textbf{36.6} & 37.1 & \textbf{39.7} & \textbf{40.0} & 38.1 & \textbf{40.1} & 37.9 & \textbf{38.8} & 36.2 & \textbf{41.1} & \textbf{37.9} & \textbf{37.7} & 36.5 & \textbf{39.4} & \textbf{38.1}\\
					Autoencoder & \textbf{27.8} & 26.4 & 27.2 & 33.0 & 33.3 & 31.8 & 29.4 & 30.9 & 30.3 & 30.8 & 28.4 & 25.1 & 26.0 & 23.7 & 23.5 & 24.7 & 24.2 & 21.8 & 23.1 & 20.3\\
					VAE         & 17.3 & 15.6 & 13.3 & 12.7 & 15.2 & 12.7 & 12.7 & 12.7 & 15.4 & 13.5 & 12.7 & 12.7 & 12.7 & 12.7 & 12.7 & 15.1 & 12.7 & 12.7 & 12.7 & 12.7\\
					PCA         & 26.0 & 25.8 & 26.5 & 32.1 & 30.2 & 31.1 & \textbf{39.4} & 38.3 & 39.0 & \textbf{40.1} & 37.5 & \textbf{38.8} & 37.4 & \textbf{36.3} & 38.0 & 37.5 & 37.0 & \textbf{39.4} & 36.3 & 36.3\\				
					\bottomrule
				\end{tabular}
			}
		\end{small}
	\end{center}
	\vskip -0.1in
\end{table}

\setlength\tabcolsep{6 pt}
\begin{table}[H]
	\caption{Residual MSEs and accuracies of various models for MNIST data and varying number of components in the outlier detection setting.  ``--'' indicates that the gradients exploded for all repetitions.}
	\label{tab:mnist_results_outlier}
	\vskip 0.1in
	\begin{center}
		\begin{small}
			\begin{tabular}{l|cccccccccc}
				\toprule
				num. comp. & 1 & 2 & 3 & 4 & 5 & 6 & 7 & 8 & 9 & 10 \\
				\midrule		
				\multicolumn{11}{l}{ResMSEs -- (i) samples of size $16\times 16$}\\		
				GT-PCA      & \textbf{0.398} & \textbf{0.315} & \textbf{0.263} & \textbf{0.228} & \textbf{0.201} & \textbf{0.180} & \textbf{0.164} & \textbf{0.150} & \textbf{0.138} & \textbf{0.128}\\
				Autoencoder & 0.690 & 0.518 & 0.452 & 0.413 & 0.387 & 0.360 & 0.332 & 0.313 & 0.300 & 0.284\\
				VAE         & 0.812 & 1.154 & -- & -- & -- & 1.163 & -- & -- & -- & -- \\
				PCA         & 0.594 & 0.541 & 0.503 & 0.476 & 0.449 & 0.426 & 0.406 & 0.388 & 0.371 & 0.356\\
				\midrule		
				\multicolumn{11}{l}{ResMSEs -- (ii) samples of size $28\times 28$}\\		
				GT-PCA      & 0.474 & 0.412 & 0.378 & 0.347 & 0.328 & 0.311 & 0.297 & 0.284 & 0.271 & 0.260\\
				Autoencoder & 0.474 & \textbf{0.393} & \textbf{0.349} & \textbf{0.302} & \textbf{0.274} & \textbf{0.250} & \textbf{0.229} & \textbf{0.209} & \textbf{0.193} & \textbf{0.176}\\
				VAE         & 0.765 & 0.402 & 1103.693 & 0.446 & 0.515 & 0.952 & 0.788 & -- & 1.149 & 1.148\\
				PCA         & \textbf{0.447} & 0.405 & 0.375 & 0.344 & 0.326 & 0.308 & 0.295 & 0.281 & 0.268 & 0.257\\
				\midrule		
				\multicolumn{11}{l}{ResMSEs -- (iii) samples of size $56\times 56$}\\		
				GT-PCA      & \textbf{0.449} & \textbf{0.371} & \textbf{0.317} & \textbf{0.279} & \textbf{0.250} & \textbf{0.228} & \textbf{0.210} & \textbf{0.194} & \textbf{0.181} & \textbf{0.170}\\
				Autoencoder & 0.761 & 0.652 & 0.498 & 0.437 & 0.409 & 0.390 & 0.359 & 0.334 & 0.313 & 0.294\\
				VAE         & 0.994 & 1.100 & 1091.176 & 1.149 & 0.985 & 3.026 & 1.098 & 0.804 & 0.816 & 0.828\\
				PCA         & 0.920 & 0.848 & 0.778 & 0.727 & 0.692 & 0.664 & 0.638 & 0.620 & 0.603 & 0.592\\
				\midrule		
				\multicolumn{11}{l}{ResMSEs -- (iv) randomly rotated samples}\\		
				GT-PCA      & \textbf{0.374} & \textbf{0.318} & \textbf{0.279} & \textbf{0.252} & \textbf{0.231} & \textbf{0.212} & \textbf{0.197} & \textbf{0.185} & \textbf{0.174} & \textbf{0.164}\\
				Autoencoder & 0.577 & 0.469 & 0.382 & 0.341 & 0.298 & 0.275 & 0.252 & 0.233 & 0.219 & 0.199\\
				VAE         & 0.738 & 0.916 & 0.643 & 0.462 & 0.383 & 0.709 & 0.872 & 0.708 & 1.411 & 164.079\\
				PCA         & 0.545 & 0.478 & 0.444 & 0.422 & 0.401 & 0.383 & 0.365 & 0.347 & 0.329 & 0.317\\
				\midrule		
				\multicolumn{11}{l}{Accuracies -- (i) samples of size $16\times 16$}\\		
				GT-PCA      & \textbf{91.2} & \textbf{91.7} & \textbf{91.8} & \textbf{91.8} & \textbf{92.2} & \textbf{92.1} & \textbf{92.3} & \textbf{92.4} & \textbf{92.2} & \textbf{92.2}\\
				Autoencoder & 90.9 & 90.8 & 90.7 & 90.8 & 89.8 & 90.5 & 86.6 & 90.8 & 88.8 & 90.5\\
				VAE         & 91.1 & 91.1 & 91.1 & 91.1 & 91.1 & 91.1 & 91.1 & 91.1 & 91.1 & 91.1\\
				PCA         & 86.7 & 90.1 & 88.3 & 88.9 & 88.7 & 88.9 & 88.8 & 88.0 & 89.0 & 88.5\\
				\midrule		
				\multicolumn{11}{l}{Accuracies -- (ii) samples of size $28\times 28$}\\		
				GT-PCA      & \textbf{91.1} & 90.6 & 90.4 & 90.2 & 90.4 & 90.6 & 90.4 & 90.5 & 90.8 & 90.9\\
				Autoencoder & 90.3 & \textbf{91.2} & 90.6 & 89.8 & 90.1 & 88.4 & 90.7 & 90.2 & 89.7 & 91.4\\
				VAE         & \textbf{91.1} & 89.8 & 90.8 & \textbf{91.1} & 91.1 & 91.2 & 91.3 & 91.1 & 91.1 & 91.1\\
				PCA         & 89.2 & 89.8 & \textbf{91.0} & 90.4 & \textbf{91.3} & \textbf{91.8} & \textbf{91.8} & \textbf{91.8} & \textbf{91.6} & \textbf{92.3}\\
				\midrule		
				\multicolumn{11}{l}{Accuracies -- (iii) samples of size $56\times 56$}\\		
				GT-PCA      & \textbf{91.1} & \textbf{91.6} & \textbf{91.8} & \textbf{91.9} & \textbf{92.1} & \textbf{92.4} & \textbf{93.2} & \textbf{93.4} & \textbf{93.5} & \textbf{93.6}\\
				Autoencoder & \textbf{91.1} & 90.8 & 87.6 & 89.7 & 89.2 & 89.2 & 89.1 & 88.3 & 90.4 & 90.7\\
				VAE         & \textbf{91.1} & 91.0 & 90.6 & 91.1 & 90.6 & 91.1 & 91.1 & 90.4 & 90.3 & 90.3\\
				PCA         & 89.9 & 89.3 & 89.4 & 87.5 & 87.4 & 85.8 & 88.0 & 87.3 & 84.6 & 88.0\\
				\midrule		
				\multicolumn{11}{l}{Accuracies -- (iv) randomly rotated samples}\\		
				GT-PCA      & \textbf{91.1} & \textbf{91.4} & \textbf{91.6} & \textbf{92.8} & \textbf{92.7} & \textbf{92.9} & \textbf{93.3} & \textbf{93.6} & \textbf{93.7} & \textbf{93.6}\\
				Autoencoder & \textbf{91.1} & 90.2 & 89.4 & 89.0 & 87.6 & 84.9 & 86.5 & 88.8 & 87.9 & 84.9\\
				VAE         & 91.0 & 91.1 & 89.6 & 89.8 & 90.7 & 90.8 & 90.3 & 91.1 & 91.1 & 91.1\\
				PCA         & 89.9 & 88.1 & 86.7 & 89.0 & 88.0 & 86.0 & 89.2 & 83.9 & 87.0 & 87.4\\
			\bottomrule
			\end{tabular}
		\end{small}
	\end{center}
	\vskip -0.1in
\end{table}

\newpage
\section{Projection of MNIST data onto GT-PCs}
\begin{figure}[ht]
	\vskip 0.1in
	\begin{center}
		\includegraphics[width=0.75\columnwidth]{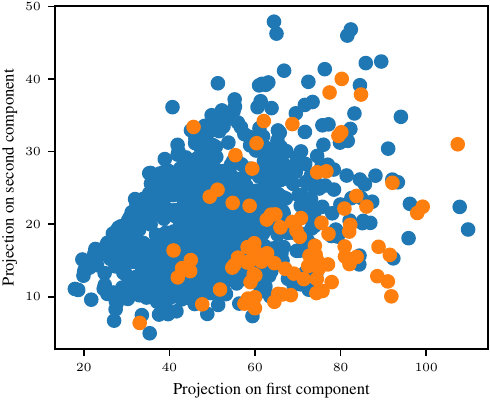}
		\caption{Projections of 1024 samples from the MNIST data set onto the first two GT-PCs in the outlier detection setting, i.\,e., fitted to images of the digit ``8''. Orange: Images of the digit ``8''. Blue: Images of other digits.}
		\label{fig:mnist_rot_proj}
	\end{center}
	\vskip -0.2in
\end{figure}

\end{document}